\documentclass{article}

\usepackage{microtype}
\usepackage{amsmath}
\usepackage{amsfonts}
\usepackage{graphicx}
\usepackage{subfigure}
\usepackage{booktabs} 
\usepackage{csquotes}
\usepackage{siunitx}
\usepackage{dcolumn}

\usepackage{hyperref}

\usepackage[unblind]{icml2018}

\usepackage{ifthen}
\usepackage{xcolor}

\newcommand*{\todo}[2][]{\textcolor{red}{[\textbf{\ifthenelse{\equal{#1}{}}{TODO}{TODO(#1)}}: #2]}}

\icmltitlerunning{On the Robustness of the CVPR 2018 White-Box Adversarial Example Defenses}

\begin{document}

\twocolumn[
  \icmltitle{On the Robustness of the CVPR 2018 White-Box Adversarial Example Defenses}



\icmlsetsymbol{equal}{*}

\begin{icmlauthorlist}
\icmlauthor{Anish Athalye}{equal,mit}
\icmlauthor{Nicholas Carlini}{equal,ucb}
\end{icmlauthorlist}

\icmlaffiliation{mit}{Massachusetts Institute of Technology}
\icmlaffiliation{ucb}{University of California, Berkeley}

\icmlcorrespondingauthor{Anish Athalye}{aathalye@mit.edu}
\icmlcorrespondingauthor{Nicholas Carlini}{npc@cs.berkeley.edu}

\icmlkeywords{Machine Learning, ICML}

\vskip 0.3in
]



\printAffiliationsAndNotice{\icmlEqualContribution} 

\begin{abstract}
    Neural networks are known to be vulnerable to adversarial examples. In this
note, we evaluate the two white-box defenses that appeared at CVPR 2018 and
find they are ineffective: when applying existing techniques, we can
reduce the accuracy of the defended models to 0\%.

\end{abstract}

\section{Introduction}
\label{sec:introduction}

Training neural networks so they will be robust to adversarial
examples~\cite{szegedy2013intriguing} is a major challenge. Two defenses that
appear at CVPR 2018 attempt to address this problem: ``Deflecting Adversarial Attacks with
Pixel Deflection''~\cite{prakash2018deflection} and ``Defense against
Adversarial Attacks Using High-Level Representation Guided
Denoiser''~\cite{liao2018denoiser}.

In this note, we show these two defenses are not effective
in the white-box threat model. 
We construct adversarial examples that reduce the classifier accuracy to
$0\%$ on
the ImageNet dataset \cite{deng2009imagenet}
when bounded by a small $\ell_\infty$ perturbation of $4/255$, a stricter
bound than considered in the original papers. Our attacks can
construct targeted adversarial examples with over $97\%$ success.

Our methods are a direct application of existing techniques.

\section{Background}
\label{sec:background}

We assume familiarity with neural networks, adversarial examples
\cite{szegedy2013intriguing},
generating strong attacks against
adversarial examples \cite{madry2018towards},
and computing adversarial examples for neural networks with non-differentiable layers~\cite{obfuscated}.
We briefly review the key details and notation.

\emph{Adversarial examples} \cite{szegedy2013intriguing}
are instances $x'$ that are very close to an
instance $x$ with respect to some distance metric ($\ell_\infty$ distance,
in this paper),
but where the classification of $x'$ is not the same as the classification of $x$.
Targeted adversarial examples are instances $x'$ whose label is equal to a
given target label $t$.

We examine two defenses: Pixel Deflection and High-level Representation Guided
Denoiser. We are grateful to the authors of these defenses
for releasing their source code and pre-trained models.

\begin{figure}
    \centering
    \includegraphics[width=\linewidth]{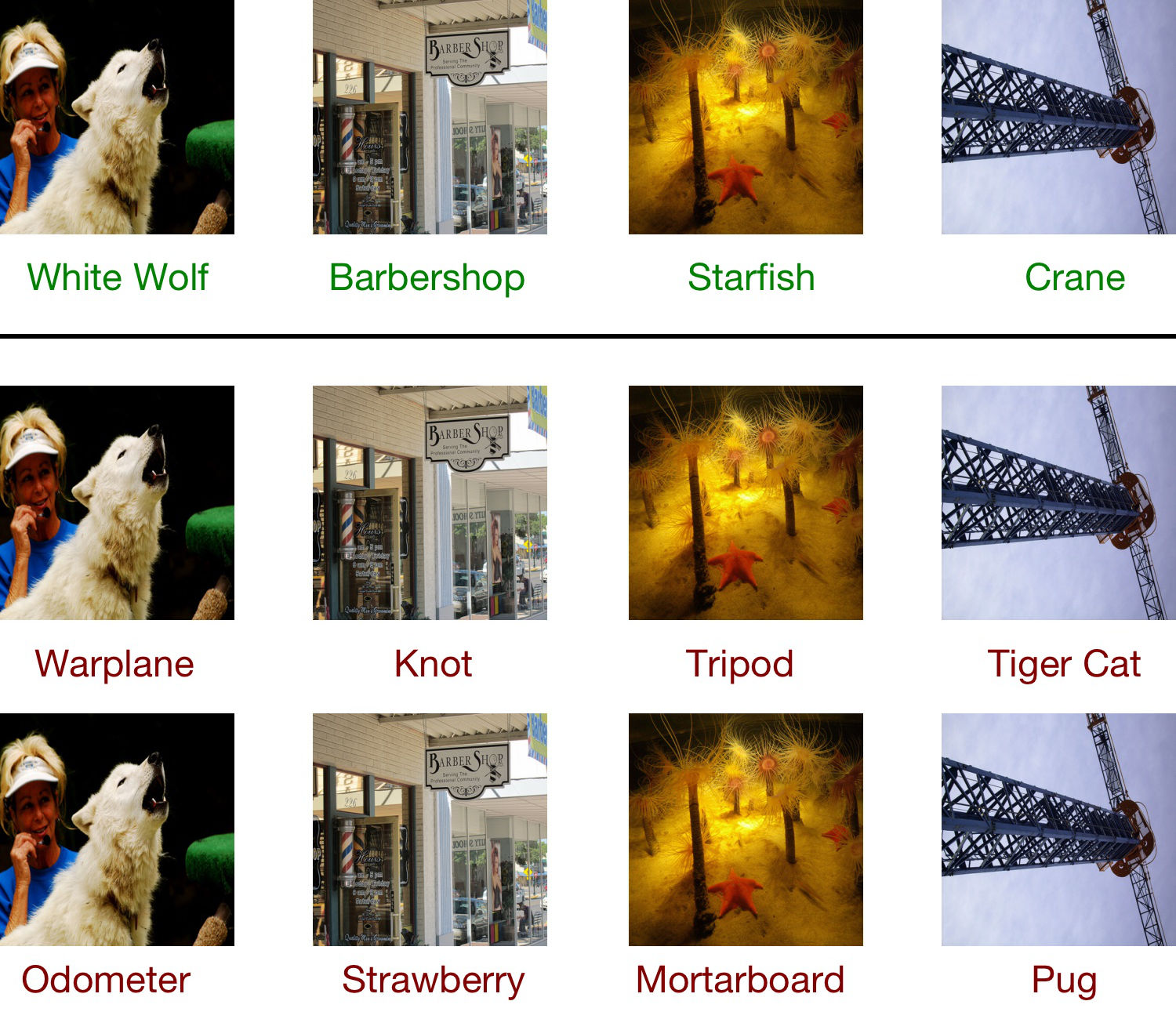}
    \vspace{-2em}
    \caption{Original images from ImageNet validation set (row 1).
      Targeted adversarial examples (with randomly chosen targets)
      for Pixel Deflection (row 2) and High-level representation Guided Denoiser (row 3),
      with a $\ell_\infty$ perturbation of $\epsilon =
    4/255$.}
    \vspace{-2em}
    \label{fig:hgd}
\end{figure}

\textbf{Pixel Deflection}~\cite{prakash2018deflection} proposes
a non-differentiable preprocessing of inputs. Some pixels (a tunable
hyperparameter) are randomly replaced with near-by pixels. This resulting
image is often noisy, and to restore accuracy, a denoising operation is
applied.

\textbf{High-level representation Guided Denoiser} (HGR)
\cite{liao2018denoiser} proposes denoising inputs using a trained
neural network before passing them to a standard classifier. This
denoiser is a differentiable, non-randomized neural network.
This defense has also been evaluated by \citet{uesato2018adversarial}
and found to be ineffective.

\subsection{Methods}
We evaluate these defenses under the white-box threat model.
We generate adversarial examples with Projected Gradient
Descent (PGD) \cite{madry2018towards} maximizing the cross-entropy loss
and bounding $\ell_\infty$ distortion by $4/255$.

\textbf{What is the right threat model
  to evaluate against?}
Many papers only
claim white-box security against
an attacker who is \emph{completely unaware} the defense is
being applied. HGD, for example, says ``the white-box attacks defined in
this paper should be called oblivious
attacks according to Carlini and Wagner's definition''
\cite{liao2018denoiser}.

Unfortunately, security against oblivious attacks is
not useful.
We only defined this threat model in our prior work
\cite{carlini2017adversarial} to study the case of an extremely
weak attacker, to show that some defenses are not even robust
under this model.
Furthermore, many
previously published schemes already achieve security
against oblivious attacks. In practice,
any serious attacker would certainly consider
the possibility that a defense is in place and try to
circumvent it, if there is a reasonable way to do so.

Thus, security against oblivious attacks is far from sufficient
to be interesting or useful in practice. Even the \emph{black-box
  threat model} allows for an attacker to be aware that the
defense is being applied, and only holds the exact parameters of
the defense as private data.
Also, our experience is that schemes that are insecure against
white-box attacks also tend to be insecure against black-box
attacks~\cite{carlini2017adversarial}.
Accordingly, in this note, we
evaluate schemes against white-box
attacks.

\section{Methodology}
\label{sec:breaks}

\subsection{Pixel Deflection}

We now show that Pixel Deflection is not robust. We
analyze the defense as implemented by the
authors~\footnote{\scriptsize{\url{https://github.com/iamaaditya/pixel-deflection}}}.
Our evaluation code is publicly
available~\footnote{\scriptsize{\url{https://github.com/carlini/pixel-deflection}}}.

We apply BPDA~\cite{obfuscated} to Pixel Deflection for
its non-differentiable replacement operation.
Our attack reduces the accuracy of the defended classifier to $0\%$.

In a targeted setting, we succeed with $97\%$ probability. (Because the
defense is randomized, we report success only if the image is classified as the
adversarial target label $9$ times out of $10$.)

\subsection{High-Level Representation Guided Denoiser}

Next, we show that using a High-level representation Guided Denoiser
is not robust in the white-box threat model.
We analyze the defense as implemented by the
authors~\footnote{\scriptsize{\url{https://github.com/lfz/Guided-Denoise}}}.
Our evaluation code is publicly
available~\footnote{\scriptsize{\url{https://github.com/anishathalye/Guided-Denoise}}}.

We apply PGD~\cite{madry2018towards} end-to-end with no modification.
It reduces the accuracy of the defended classifier to $0\%$ and
achieves $100\%$ success at generating targeted adversarial examples.


\section{Conclusion}
\label{sec:conclusion}

As this note demonstrates, Pixel Deflection and High-level
representation Guided Denoiser (HGD) are not robust to adversarial examples.

\section*{Acknowledgements}

We are grateful to Aleksander Madry and David Wagner
for comments on an early draft of this paper.

We thank Aaditya Prakash and Fangzhou Liao for discussing their defenses
with us, and we thank the authors of both papers for releasing source code and
pre-trained models.

\bibliography{paper}
\bibliographystyle{icml2018}

\end{document}